%% file: mugen.tex
\documentclass[10pt,journal]{IEEEtran}

\usepackage{amssymb}
\usepackage{graphicx}
\usepackage{amsmath,bm}
\usepackage{tabularx}  
\usepackage{times}
\usepackage{multicol}
\usepackage[bookmarks=true]{hyperref}
\usepackage{xspace}
\usepackage{booktabs}
\usepackage{wrapfig}
\usepackage{listings}
\usepackage[dvipsnames]{xcolor}
\usepackage{utfsym}
\usepackage{colortbl}
\usepackage{caption}
\usepackage{arydshln}
\usepackage[ruled]{algorithm2e}
\usepackage{multirow}
\usepackage{subcaption}

\newcommand{\encoder}{\ensuremath{\mathcal{E}}}
\newcommand{\embedding}{\ensuremath{\texttt{Emb}}}
\newcommand{\decoder}{\ensuremath{\mathcal{D}}}
\newcommand{\codebook}{\ensuremath{\mathcal{B}}}

\newcommand{\stt}{\ensuremath{{s}}}
\newcommand{\act}{\ensuremath{{a}}}
\newcommand{\refm}{\ensuremath{{m}}}
\newcommand{\policy}{\ensuremath{{\pi}}}
\newcommand{\latent}{\ensuremath{{z}}}
\newcommand{\wm}{\mathcal{{W}}}

\newcommand{\loss}{\ensuremath{\mathcal{L}}}

\newcommand{\simBuffer}{\ensuremath{\texttt{Buff}}}

\newcommand{\fig}{Figure{}~}
\newcommand{\eqn}{Equation{}~}
\newcommand{\Sec}{Section{}~}
\newcommand{\Tab}{Table{}~}
\newcommand{\alg}{Algorithm{}~}
\newcommand{\app}{Appendix{}~}

\title{\LARGE \bf
MuGen: Multi-Skill Generative Locomotion Controller for Humanoid Robots
}

\author{
  Yusen Feng$^{*}$,
  Xiang Wang$^{*}$,
  Heyuan Yao,
  Zixi Kang\\
  Xinyu Huo,
  Boyang Yu,
  Pengyun Qiu,
  Ruijie Zhao\\
  Baoquan Chen$^{\dagger}$,
  Libin Liu$^{\dagger}$ %
  \thanks{$^{*}$Equal contribution.}
  \thanks{$^{\dagger}$Corresponding authors.}%
}
\IEEEspecialpapernotice{{Peking University}}

\begin{document}
\maketitle

\begin{figure*}[t]
    \centering
    \vspace{-30pt}
    \includegraphics[width=1\linewidth]{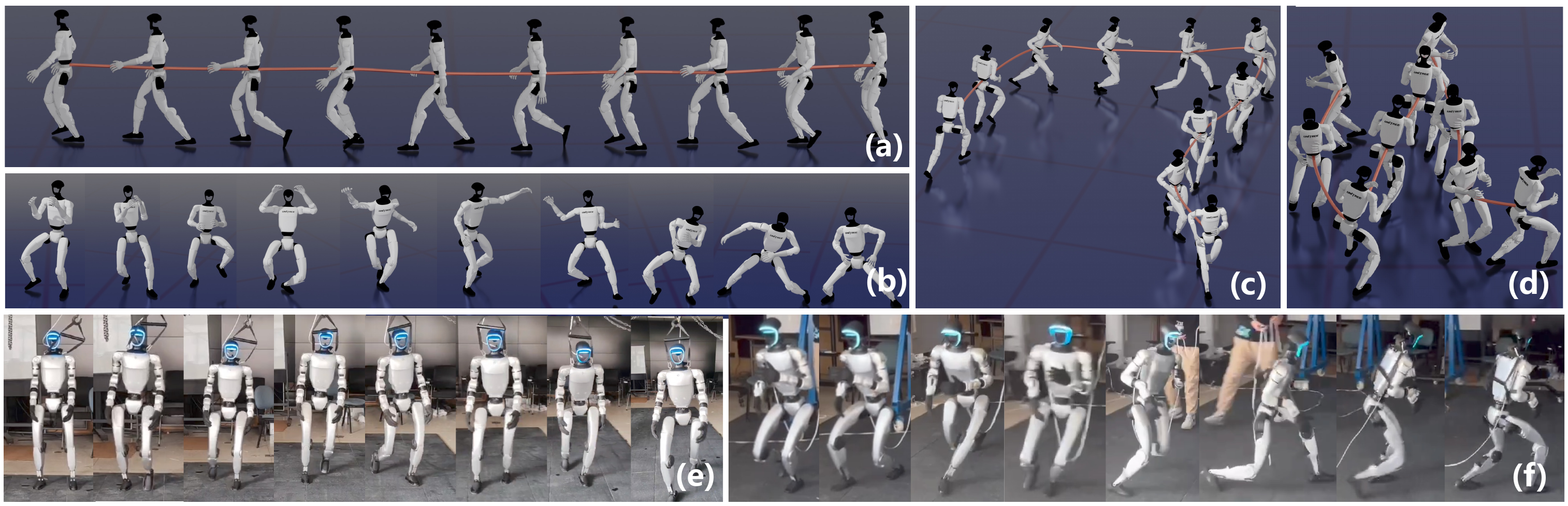}
    \caption{\textbf{MuGen} enables multi-skill humanoid locomotion by learning a generative controller.
    {(a-d)}: A simulated humanoid tracks (a) a long walking motion manually crafted from motion data and (b–d) dance, run, and crouching walk motions selected from the motion dataset.
    {(e-f):} A real Unitree G1 robot tracks short segments of (e) a straight walk and (f) a crouching walk from the motion dataset. 
    Details are provided in \Sec\ref{sec:simu_results}.
    }
    \label{fig:teaser}
    \vspace{-0.4cm}
\end{figure*}

\begin{abstract}

This paper presents MuGen, a data-driven framework for learning and deploying multi-skill locomotion on humanoid robots. MuGen enables a robot to perform expressive motions like humans under the guidance of example motion sequences. To achieve this, we employ vector-quantized autoencoders (VQ-VAEs) trained with model-based reinforcement learning, resulting in a generative representation of locomotion that captures key patterns of human motion from hours of heterogeneous human performance data. We employ a teacher-student learning framework and develop a new policy distillation strategy to enable a deployable student policy learning this efficient latent representation. This policy allows the robot to track and mimic unseen human motions and further enables the robot to reuse the learned latent space for other tasks. We demonstrate the effectiveness of our framework through a diverse set of motions and accurate execution.
\end{abstract}


\input{sections/1_introduction}

\section{Related Works}

\subsection{Legged Robot Locomotion}
Legged locomotion has long been a fundamental topic in robotics, with early approaches primarily centered on model‑based control and trajectory optimization.  
Classical methods, ranging from the pioneering WABOT~\cite{kato1973development} and early dynamic walking studies~\cite{miura1984dynamic} to the landmark Honda humanoids~\cite{hirai1998development} rely heavily on accurate analytic models~\cite{8461207,ramos2019dynamic,MontecilloPuente2010OnRW,6907261,7258356}.  
Subsequent work introduced hybrid‑zero dynamics~\cite{westervelt2003hybrid}, whole‑body task hierarchies~\cite{dariush2008whole}, and the 3‑D LIPM for pattern generation~\cite{kajita20013d}.  
Model‑based techniques remain state‑of‑the‑art for many humanoid and quadruped platforms, including ANYmal~\cite{hutter2016anymal}, geometric retargeting~\cite{darvish2019wholebody}, multimode or tele‑operated frameworks~\cite{penco2019multimode,joao2019dynamic, 9158331, 10380694}, whole‑body MPC~\cite{moro2019whole, romualdi2022online,englsberger2020mptc,elobaid2023online}, and recent acrobatic or synchronized human–robot studies~\cite{chignoli2021humanoid,dallard2023Sync}.  
Despite their effectiveness, these methods demand extensive expert tuning and often degrade under modeling error or unstructured terrain.

\subsection{Reinforcement Learning-based Methods for Locomotion}
Recent advances in reinforcement learning (RL) have shown remarkable robustness and adaptability, as policies are learned directly from interaction without an exact dynamics model.  
For quadrupeds, model‑free RL enables agile motion imitation~\cite{Peng2020-ty}, rapid motor adaptation~\cite{kumar2021rma}, energy‑efficient gait discovery~\cite{fu2021minimizing}, vision‑based blind stair climbing~\cite{siekmann2021blind}, and parkour‑level versatility~\cite{zhuang2023robot,cheng2023parkour,wang2025beamdojo}.  
RL further supports loco‑manipulation~\cite{cheng2023legmanip,jeon2023learning,schwarke2023curiosity} and dynamic ball‑dribbling~\cite{ji2023dribblebot}.  
For bipeds and humanoids, policies have progressed from simulation‑only agility~\cite{peng2018deepmimic, 2021-TOG-AMP, 2022-TOG-ASE, tessler2023calm} to real‑world deployment~\cite{li2021reinforcement,li2024reinforcement, RealHumanoid2023}, RL exoskeleton‑cockpit teleoperation~\cite{ben2025homie}
, and versatile jump/stand‑up controllers~\cite{li2023robust,huang2025learning}.  
Nevertheless, low sample efficiency and the sim‑to‑real gap remain key obstacles.  Some works ~\cite{simgan2021, he2025asap} combine online system identification with policy learning, while extensive domain randomization is still common practice~\cite{agarwal2023legged,yang2023neural,duan2023learning}. However, these tasks often require fine-grained reward design with respect to the goal. In this work, we address these limitations by proposing a differentiable physical simulator, transforming humanoid behaviors learning into a model-based approach. Our method reduces the need for extensive reward engineering and enables direct transfer from simulation to real robots.

\subsection{Imitation Learning‑based Methods for Locomotion}
Imitation learning (IL) leverages motion‑capture or video demonstrations to mitigate reward‑design burdens. Recent work in physics‑based motion control condenses large‑scale human‑motion datasets into compact skill sets via generative controllers. This  began with VAE models~\cite{ling2020character, physics22vae}, followed by adversarial motion priors~\cite{ho2016generative, peng2021amp}, evolving toward reusable skill embeddings~\cite{2022-TOG-ASE, yao2022controlvae, tessler2023calm, zhu2023neuralcategoricalpriorsphysicsbased, yao2024moconvq} and masked‑motion inpainting~\cite{tessler2024maskedmimic}. As for the deployment, 
{Quadruped IL} combines trajectory optimization with policy distillation~\cite{fuchioka2023opt, huang2024diffuseloco}, adversarial priors in the wild~\cite{Escontrela22arXiv_AMP_in_real,wang2023amp}, model-based learning with VAE
~\cite{shi2024efficientmodelbasedapproachlearning}, and multi‑modal whole‑body control~\cite{dugar2024mhc}.  
{Humanoid IL} explores expressive whole‑body control~\cite{cheng2024exbody,ji2024exbody2}, omni‑directional tele‑operation~\cite{he2024omnih2o, he2024learning,he2024hover}, instant human camera data~\cite{fu2024humanplus}, next‑token transformer control~\cite{TokenHumanoid2024}, universal motion representations~\cite{luo2024universal}, generative strategy learning~\cite{wang2024strategy}, surrogate function tuning~\cite{robotmdm24}, and GAN‑based style distillation~\cite{ma2025styleloco}.  
While these combinations greatly improve motion realism and sample efficiency, reliable transfer to hardware still hinges on accurate dynamics alignment and sensor calibration. Potential solutions include AMP‑in‑real adaptations~\cite{RoboImitationPeng20,he2024hover} and using delta action model that compensates for the
dynamics mismatch~\cite{he2025asap}. Motivated by VAE based works~\cite{yao2022controlvae, shi2024efficientmodelbasedapproachlearning, zhu2023neuralcategoricalpriorsphysicsbased, yao2024moconvq}, our approach complements these efforts by introducing a VQVAE encoder–decoder that learns a discrete, dynamics‑feasible motion latent space from heterogeneous human data, enabling zero‑shot composition and real‑world execution of diverse humanoid skills.

\section{Method}
\label{sec:method}

\begin{figure*}[t]
    \centering
    \includegraphics[width=0.95\linewidth]{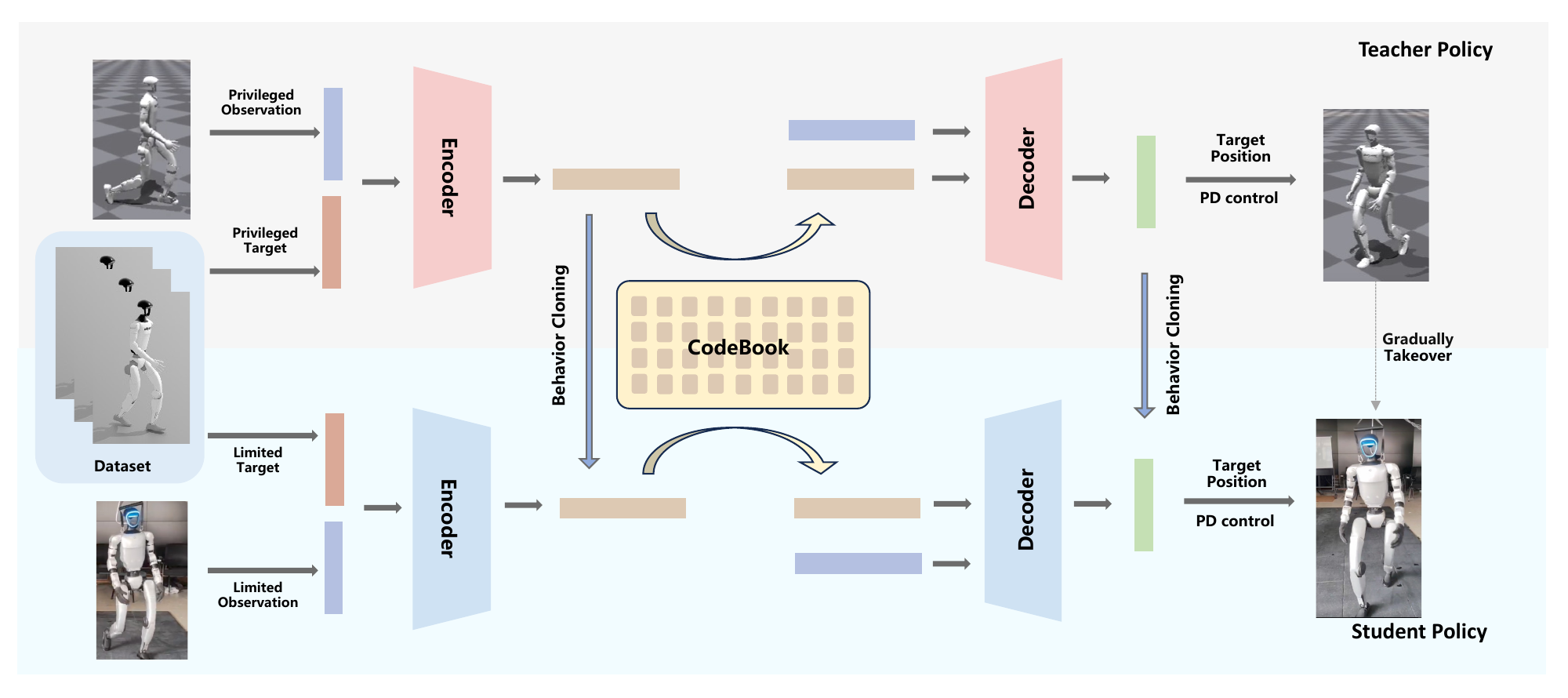}
    \caption{\textbf{System overview}  
    1) \emph{Motion Skill Embedding:} states and reference motions are encoded into continuous representations, then a VQ bottleneck maps embeddings to a trainable codebook.   
    2) \emph{Student policy:} using the shared codebook, the student decodes actions from partial observations and aligns them with teacher outputs through behavior cloning.  
    All training is performed in simulation environment.
    }
    \label{fig:pipeline}
\end{figure*}

We propose MuGen, a data-driven framework for learning and deploying multi-skill locomotion on humanoid robots. As illustrated in Figure~\ref{fig:pipeline}, MuGen employs a teacher-student architecture for robust motor skill acquisition. The teacher learns skills using privileged information, aided by a world model that enhances sample efficiency and enables efficient optimization via backpropagation. The student distills these skills for deployment with limited information. Skill knowledge is transferred using a shared Vector-Quantized (VQ) codebook as a skill space, combined with simultaneous behavior cloning in both this VQ space and the raw action space. We elaborate on these components in the following sections.

\subsection{Problem Formulation}
We target generative humanoid locomotion by training controllers using model-based reinforcement learning (MBRL). The task is framed as a partially observable Markov decision process (POMDP), represented by the tuple $\mathcal{M}= \langle\mathcal{S},\mathcal{O},\mathcal{G},\mathcal{A},\mathcal{P},\mathcal{R}\rangle$. Here, $\mathcal{S}$, $\mathcal{O}$, $\mathcal{G}$, and $\mathcal{A}$ denote the state, observation function, goal, and action spaces, respectively. $\mathcal{P}(s'\!\mid\!s,a)$ describes the true environment transition dynamics, and $\mathcal{R}(s, a, g)$ is the reward function. In our locomotion context, $\mathcal{R}$ typically quantifies how closely the robot's state or resulting motion aligns with reference motion capture data, as specified by the goal $g$. The standard objective in this framework is to find a policy $\pi$ that maximizes the expected discounted cumulative reward $\mathbb{E}[\sum_{t=0}^{\infty} \gamma^t \mathcal{R}_t]$.

Instead of maximizing returns in the true environment $\mathcal{P}$ (often requiring sample-based gradients), our MBRL approach employs a learned, differentiable world model $\mathcal{W}(s'\!\mid\!s,a)$, which approximates the true dynamics $\mathcal{P}$. The objective function turns to: 
\begin{equation}
    \min_{\theta}\; \mathbb{E}_{\substack{s_{0}\sim\rho(s_0),\,a_{t}\sim\pi_\theta(\cdot|s_t),\\s_{t+1}\sim\mathcal{W}(\cdot\mid s_t,a_t)}}\!\biggl[\sum_{t=0}^{\infty}\gamma^t\,\mathcal{L}\bigl(s_t,a_t,g_t\bigr)\biggr],
\label{eq:policy_loss}
\end{equation}
where $\policy_\theta(\act_t\mid \stt_t)$ is a parameterized control policy, $\gamma\in(0,1]$ is the discount factor and $\mathcal{L}$ is a differentiable loss with $\mathcal{L} = -\mathcal{R}$. The outer expectation involves sampling actions from the policy $\pi_\theta$ and predicting next states $s_{t+1}$ using the differentiable world model $\mathcal{W}$. Because both $\pi_\theta$ and $\mathcal{W}$ are differentiable, we can efficiently compute the gradients of the loss objective with respect to the policy parameters $\theta$ using backpropagation through the predicted state trajectories. This avoids the need for high-variance, sample-based gradient estimators often used in model-free RL and enables stable, end-to-end gradient-based optimization of the control policy.

\subsection{Dynamics Model}
We adopt a world model that serves as the predictor of the next state for environment modeling. At time step $t$, the world model takes the current robot state $s_t$ and the action $a_t$ produced by the teacher policy, and predicts the change of state between time steps. 
We then integrate the change with the current robot state to obtain the predicted next state $s_{t+1}$.
The world model is optimized by minimizing reconstruction loss:
\begin{equation}
    \mathcal{L}_{\mathcal{W}} = \sum_{t=0}^{T-1} \gamma^t\bigl\| W_{\mathcal{W}} (s_{t+1}^{\text{sim}} - s_{t+1}^{\mathcal{W}}) \bigr\|_1,
    \label{eq:world_model_loss}
\end{equation}
where \(W_{\mathcal{W}}\) represents a predefined set of weights the metric.
$\mathcal{L}_{\mathcal{W}}$ penalizes deviations between the simulated next states \(s_{t+1}^{\text{sim}}\) and the world model's predictions \(s_{t+1}^{\mathcal{W}}\).
Notably, to build a more precise prediction, the world model has access to full privileged information as input. Therefore, we jointly optimize the teacher policy and the world model through alternating updates. During the distillation process, we freeze the world model to prevent performance degradation when we begin the distillation process.

\subsection{Skill Embeddings}

As shown in \fig~\ref{fig:pipeline}, the teacher policy comprises an encoder $\encoder$ and a decoder $\decoder$. The encoder $\encoder$ maps states $s$ and reference motion $m$ to continuous embeddings $z = \encoder(s, m)$, which are then quantized by a VQ bottleneck to the nearest codebook entry $\hat{z} = \codebook(z)$. The decoder $\decoder$ conditions on the code and the current state to output a distribution over joint targets, $a \sim \decoder(\cdot|\hat{z}, s)$, encouraging reuse of motion primitives and stabilizing exploration. Since the simulation environment is not differentiable, we adopt world model to predict the resulting next state $s_{t+1}^{\mathcal{W}}\sim\mathcal{W}(\cdot|s,a)$ for gradient backpropagation.
We jointly optimize the teacher policy and the codebook during the same update steps with the objective function:
\begin{align}
    \mathcal{L}_{T} = & \sum_{t=0}^{T-1}\gamma^t\biggl(\underbrace{ \bigl\| W_{\mathcal{T}} (s^{\mathcal{W}}_{t+1} - m_{t+1}) \bigr\|_1}_{\mathcal{L}_{\text{rec}}} \nonumber\\
    &+ \underbrace{\beta_1 \bigl\|\operatorname{sg}(\encoder(s_t,m_t)) - \hat{z}_t\bigr\|_2}_{\mathcal{L}_{\text{VQ}}} \nonumber\\
    &+\underbrace{\beta_2 \bigl\|a_t\bigr\|_2+\beta_3 \bigl\|a_{t}-a_{t-1}\bigr\|_2}_{\mathcal{L}_{\text{reg}}}\biggr),
    \label{eq:teacher_loss}
\end{align}
where \(W_{ {T}}\) denotes weights of state-component which will be shown in \Sec\ref{app:obs_space}, $\beta_1$ controlling the commitment weights. 
$\mathcal{L}_{\text{rec}}$ enforces consistency between reconstructed and reference states, and $\mathcal{L}_{\text{VQ}}$
regularizes the latent space using stop-gradient operators $\operatorname{sg}(\cdot)$. Following the approach suggested in the appendix of \cite{vqvae}, we replace the optimization stage of codebook update loss by applying the strategy of exponential moving averages (EMA) to update the dictionary items.
The whole objective function $\mathcal{L}_{T}$ simultaneously enforces both behavioral fidelity and latent space regularity. We further apply two regularization terms $\mathcal{L}_{\text{reg}}$ on the decoded action to encourage smooth and normal actions. 
It is notable that the codebook will only be updated during this stage, and will remain frozen during the training of the student policy.

\subsection{Policy Distillation}

The teacher-student framework is widely used for deployable policy distillation \cite{he2024omnih2o, ji2024exbody2, wu2025learnteachsampleefficientprivileged, wang2024ctsconcurrentteacherstudentreinforcement}. However, this process often encounters a capability drop due to the lack of privileged information, which is only available in simulation and significantly enhances policy performance. To bridge this gap, we propose teacher-student architecture with distinct observation space\footnote{\app \ref{app:obs_space} will show detailed observation space for both teacher and student policies.} but a shared latent skill domain. 
The key component of the generative policy, the VQ codebook \(\codebook\), is shared between the teacher and student. \(\codebook\) enables the student to leverage high-quality dynamics and motion skills learned by the teacher, thereby implicitly transferring knowledge accessible only through privileged information.
The behavior cloning objective function of student policy combines action-level and latent-space matching:
\begin{align}
    \mathcal{L}_{S} &= \sum_{t=0}^{T-1}\gamma^t\biggl(\underbrace{\left[\|a^{\mathrm{S}}_t - a^{\mathrm{T}}_t\|_2\right]}_{\mathcal{L}_{\text{BC}}} \nonumber\\
    &+ \beta_{4} \underbrace{\mathbb{E}_{s_t \sim \rho_{\pi^{\mathrm{S}}}}\left[\|\encoder^{\mathrm{S}}(s_t, g_t) - \encoder^{\mathrm{T}}(s_t^{full}, m_t)\|_2\right]}_{\mathcal{L}_{\text{align}}}\biggr),
    \label{eq:student_loss}
\end{align}
where \(\rho_{\pi^{\mathrm{S}}}\) denotes the state space observed by student policy and \(\beta_{4}\) is a balancing weight.
$\mathcal{L}_{\text{align}}$ encourages the student encoder \(\encoder^{\mathrm{S}}\) to produce embeddings compatible with the teacher's frozen codebook \(\codebook\), enabling feature-level knowledge transfer while maintaining flexibility. 
 $\mathcal{L}_{\text{BC}}$ is a standard behavior cloning loss
which directly minimizes the action discrepancy between the student and teacher.

\subsection{Training Schedule}
\label{sec:schedule}
The training process of MuGen is divided into three main stages: (1) \emph{Pretraining}: build the skill latent space within the VQ codebook by training teacher policy. (2) \emph{Warming up}: refine the teacher policy and codebook, and warm up the student policy. (3) \emph{Distillation}: allow the student policy to gradually take over control to obtain a deployable policy. 

During the \emph{Pretraining} stage, we optimize and roll out only the teacher policy. The teacher policy and world model are jointly trained through alternating updates using \eqn\eqref{eq:world_model_loss} and \eqn\eqref{eq:teacher_loss}. We do not optimize the student policy at this stage because the teacher is still exploring and building the codebook, and an unstable target policy could harm early student learning.

In the \emph{Warming Up} stage, we optimize all modules, include the student policy, while continuing to roll out trajectories using only the teacher policy. Once the teacher has sufficiently learned and student has warmed up, we proceed to the distillation stage, where the student policy gradually takes over robot control.

In the \emph{Distillation} stage, the teacher policy and the codebook are frozen, and the student policy is continually trained to distill the knowledge of the teacher. We establish a smooth transition between the teacher and student: in each rollout trajectory of this stage, we randomly choose between \(\pi^{\mathrm{T}}\) and \(\pi^{\mathrm{S}}\) at every step, with the probability of selecting the teacher policy linearly annealed to zero. This strategy differs from the original  DAgger~\cite{dagger}, as the policies are mixed within a single trajectory, allowing the teacher to stabilize the robot when the student makes incorrect decisions.
This linear annealing strategy mitigates abrupt distribution shifts while preserving the benefits of teacher oversight during early training stages. 

After completing the \emph{Distillation} stage, MuGen is ready for deployment. For more details, please refer to the pseudocode provided in \alg\ref{alg:training_mugen} in the Appendix.

\begin{figure*}[!t]
    \centering
    \includegraphics[width=0.7\linewidth]{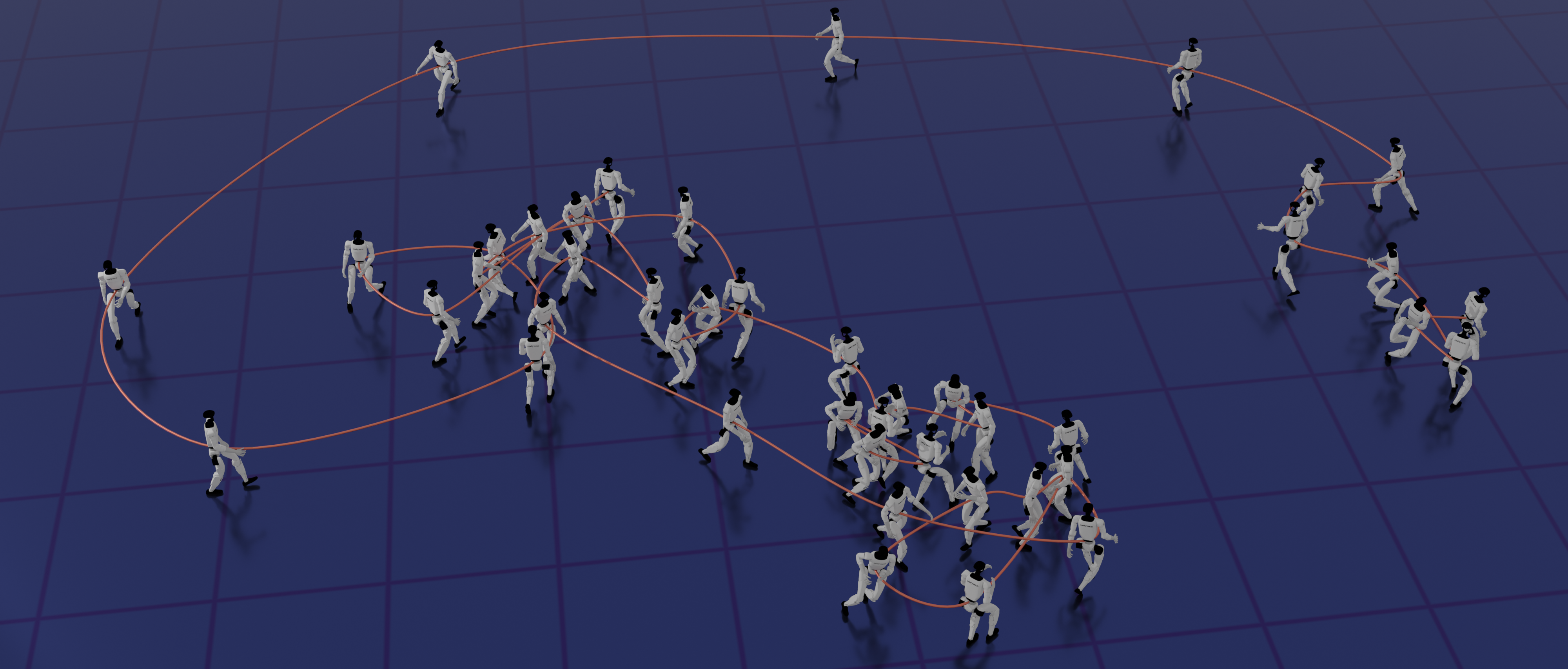}
    \caption{The trajectory generated by the student policy using the learned prior encoder $\hat{\encoder}=\hat{\encoder}(z|s)$. }
    \label{fig:sample}
    \vspace{-0.4cm}
\end{figure*}

\section{Experimental Results}
\label{sec:result}

Our training and evaluation framework is implemented using the NVIDIA IsaacGym~\cite{makoviychuk2021isaac} simulator.
To bridge the sim-to-real gap, we apply domain randomization during training by varying simulation parameters, adding random noise to observations, and applying random weight loads to the robot’s torso. For more details, please refer to Appendix~\ref{sec:domain_randomization}.
We utilize the retargeted version of the LaFAN1 motion dataset~\cite{harvey2020robustLafanYHY} provided by Unitree~\cite{unitreerobotics_huggingface2025} for training and testing MuGen. Specifically, we use a selected subset of this dataset containing only locomotion, with a total duration of roughly one hour. Appendix~\ref{sec:dataset} provides detailed statistics of this locomotion dataset.

\subsection{Simulation Results}\label{sec:simu_results}
We evaluate MuGen’s performance in simulation using the one-hour selected subset of the LaFAN1 dataset. As illustrated in \fig\ref{fig:teaser}, MuGen’s student policy successfully tracks various motions using only deployable observations. Specifically, \fig\ref{fig:teaser}(c) and (d) show results for tracking a motion sequence from the same dataset. The reference motion in \fig\ref{fig:teaser}(a) is created by repeating the kinematic trajectory of a walking cycle selected from the dataset. We find that the MuGen policy can maintain balance even when tracking unseen motions that are very different from the training data, such as a dance. However, the robot struggles to perform such motions using only the learned skills from the training set. To demonstrate that our framework is capable of robustly controlling more dynamic skills, we train a new MuGen policy using a single reference dance clip. The student policy can then perform the dance accurately, as shown in \fig\ref{fig:teaser}(b).
These results confirm that MuGen can generalize well to diverse locomotion tasks when trained on unstructured data and achieve precise tracking performance under targeted training.

\subsubsection{Generation Ability}
We further evaluate the generative ability of our deployable student policy and design an experiment in which the student autonomously produces motion by sampling latent skills from the pretrained VQ codebook. This is carried out in a post-training stage, where we introduce a new encoder called prior encoder $\hat{\encoder}$, which only accepts the state as input, while freezing all other components.
The prior encoder $\hat{\encoder}=\hat{\encoder}(z|s)$ takes only proprioception input and outputs a probability distribution over discrete codebook indices. A corresponding latent skill embedding $\hat{z}$ is then extracted from the codebook and passed into the frozen decoder $\decoder$. This setup allows us to generate actions directly from onboard observations using learned skills.

To train $\hat{\encoder}$, we let the teacher policy track randomly selected reference motions and treat the resulting codebook indices as labels. The prior is trained via a standard cross-entropy loss to predict these indices. During inference, we compose a fully deployable policy by combining the trained prior encoder $\hat{\encoder}$, the codebook $\codebook$, and the student decoder $\decoder$.
\fig\ref{fig:sample} demonstrates that the student policy, conditioned only on the proprioceptive state, can generate smooth and diverse behaviors, transitioning between different motion patterns such as walking, running, crouching, and turning around, without any reference motion. These results validate the expressiveness and robustness of MuGen's learned skill space and its capacity for real-time generative control.

\begin{table*}[t]
\centering
\caption{Evaluation on tracking task for MuGen and baselines.}
\label{tab:baseline}
\resizebox{1.\textwidth}{!}{
    \begin{tabular}{ccccccc}
    \toprule
    Model & SR$_{seen}$$\uparrow$ & MJRE$_{seen}$ (rad)$\downarrow$ & MVE$_{seen}$$\downarrow$ & SR$_{unseen}$$\uparrow$ & MJRE$_{unseen}$ (rad)$\downarrow$ & MVE$_{unseen}$$\downarrow$ \\
    \midrule
    baseline w/o history & \textbf{0.9082} & \textbf{0.0759} & 0.3581 & 0.6070 & 0.0996 & 0.4832 \\
    VAE-based w/o history & 0.6490 & 0.0842 & 0.3519 & 0.5151 & 0.1094 & 0.4359 \\
    VQ-based w/o history & 0.8072 & 0.0977 & 0.4303 & 0.7279 & 0.1108 & 0.4686 \\
    MuGen(Ours) & 0.8820 & 0.0791 & \textbf{0.3404} & \textbf{0.9535} & \textbf{0.0927} & \textbf{0.3652} \\
    \bottomrule
    \end{tabular}%
}
\end{table*}

\begin{table*}[t]
\centering
\caption{Ablation of History Observation.}
\label{tab:ablation}
\resizebox{0.9\textwidth}{!}{
    \begin{tabular}{cccccccc}
    \toprule
    \#HT & \#HS & SR$_{seen}$$\uparrow$ & MJRE$_{seen}$ (rad)$\downarrow$ & MVE$_{seen}$$\downarrow$ & SR$_{unseen}$$\uparrow$ & MJRE$_{unseen}$ (rad)$\downarrow$ & MVE$_{unseen}$$\downarrow$ \\
    \midrule
    0 & 0 & 0.8072 & 0.0977 & 0.4303 & 0.7279 & 0.1108 & 0.4686 \\
    0 & 5 & \textbf{0.8820} & \textbf{0.0791} & \textbf{0.3404} & \textbf{0.9535} & \textbf{0.0927} & \textbf{0.3652} \\
    2 & 5 & 0.8666 & 0.0907 & 0.3674 & 0.6810 & 0.1101 & 0.4230 \\
    5 & 5 & 0.2006 & 0.0925 & 0.4455 & 0.2615 & 0.1170 & 0.4888 \\
    2 & 10 & 0.6901 & 0.0824 & 0.3605 & 0.4824 & 0.1066 & 0.4156 \\
    \bottomrule
    \end{tabular}%
}
\end{table*}

\subsection{Ablation Study}\label{sec:ablation}

To assess the key components of our system design, we conduct ablation studies that evaluate the performance of policies under various settings. The goal is to measure how well each policy can maintain stable tracking under random perturbations introduced by domain randomization.
We compare MuGen against two baselines: (1) a vanilla MLP policy that directly maps input states and reference motions to actions, and (2) a VAE-based policy, where the latent skill space is continuous rather than discretized. Details of these baselines can be found in Appendix~\ref{sec:ablation_baseline}. All models are trained on a small subset of the LaFAN1 dataset consisting of approximately 200 seconds of walking motions and tested on both the same motion set and an additional test subset of approximately 90 seconds of walking motions. Detailed information about the dataset used for the ablation study is provided in Appendix~\ref{sec:dataset}. 

We evaluate the performance of these policies by tracking the same reference motions in 1,024 simulation environments, each initialized with different domain randomization parameters. To assess robustness, we report the following metrics:
(1) \textbf{Survival Rate (SR)}, the proportion of environments completed without falling, 
(2) \textbf{Mean Joint Rotation Error (MJRE)}, the average local rotation error between each joint and its reference joint over successful rollouts, measured in radians, and
(3) \textbf{Mean linear Velocity Error (MVE)},the average deviation of root velocity from the reference.
Results are summarized in \Tab\ref{tab:baseline}. 
MuGen achieves strong performance, with higher SR and lower MJRE on both seen and unseen motions, demonstrating better generalization and greater resilience to disturbances compared to both baselines. The vanilla MLP policy achieves strong performance on the training dataset but shows a dramatic performance drop on unseen data, indicating overfitting to the training set. These findings highlight the effectiveness of discrete skill embeddings in enhancing policy robustness.

\begin{figure}[t]
  \centering
  \begin{subfigure}[b]{0.4\textwidth}
    \centering
    \includegraphics[width=\textwidth]{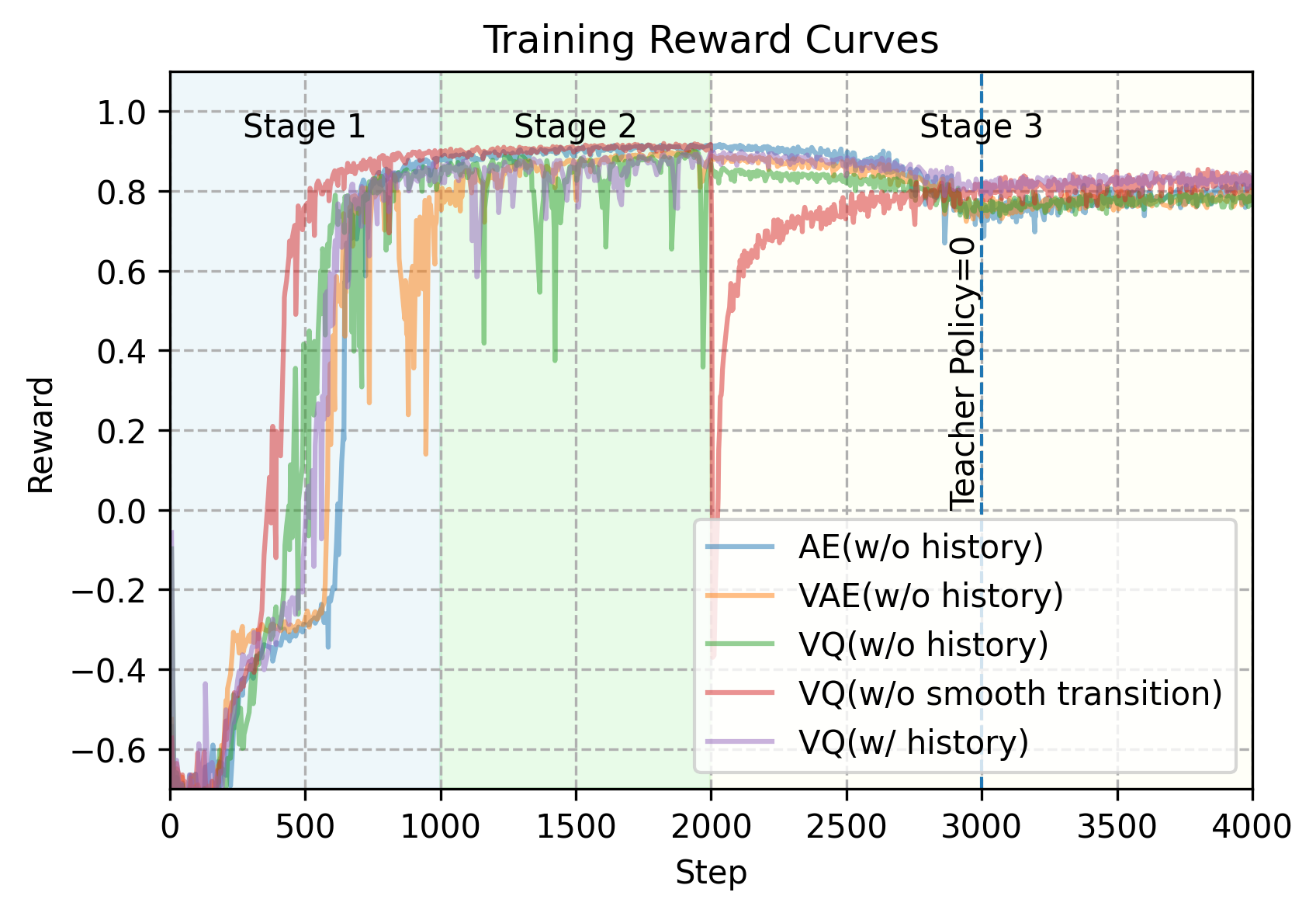}
    \label{fig:figure1}
  \end{subfigure}
  \begin{subfigure}[b]{0.4\textwidth}
    \centering
    \includegraphics[width=\textwidth]{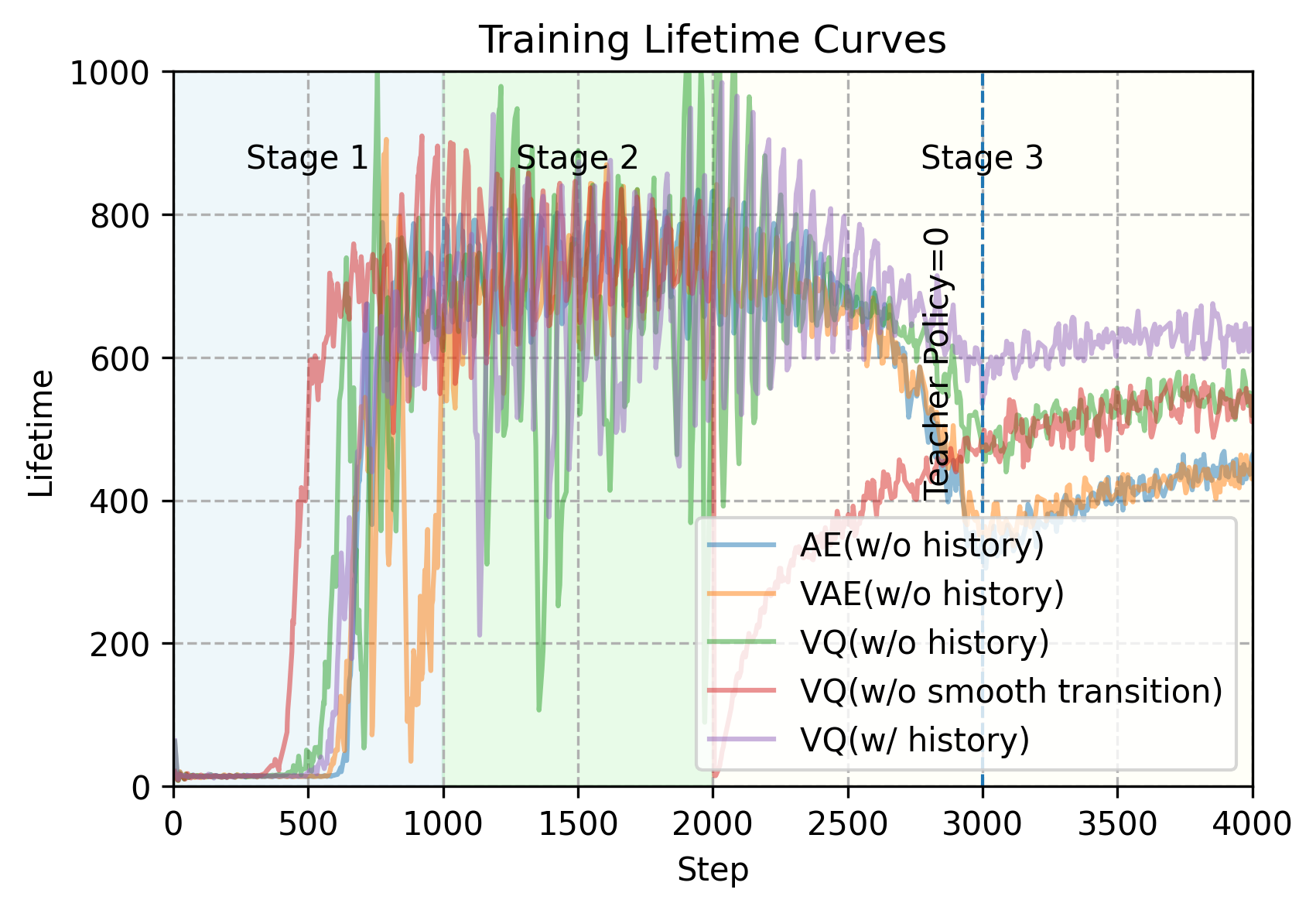}
    \label{fig:figure2}
  \end{subfigure}
  \caption{Training Curves of different model designs. We calculate weighted tracking error and map it into reward matrix with $e^{-wx}$, where $w$ is a weight factor and $x$ is the tracking error in global space. }
  \label{fig:side_by_side}
\end{figure}

We further ablate the importance of the length of temporal context in the state representation and the smooth transition strategy described in \Sec\ref{sec:schedule}. 
\textbf{Temporal Context:} We vary the number of historical frames observed by the teacher (\#HT) and student (\#HS) policies to assess the role of temporal information. Results in \Tab\ref{tab:ablation} indicate that incorporating a moderate history horizon improves tracking accuracy and survival rate by providing better motion continuity and contextual cues. However, using excessive history introduces noise and may exceed the representational capacity of the policy networks, leading to reduced performance. 
\textbf{Smooth Transition:} As illustrated in \fig\ref{fig:side_by_side}, disabling the smooth transition mechanism, i.e., switching control from the teacher to the student at the beginning of the \emph{Distillation} stage, leads to a significant performance drop. The training becomes unstable during the stage-switching period, which in turn degrades the final  performance. This demonstrates the importance of our scheduling strategy in policy distillation.

\subsection{Real-World Deployment}
Our experiments are conducted on the Unitree G1 humanoid robot (35\,kg, 1.32\,m height) equipped with 23 actuated degrees of freedom (legs: 6\,$\times$\,2, arms: 5\,$\times$\,2, waist: 1).  
The policy is executed at 30\,Hz to generate target joint position and a low-level PD controller is actuated at 500\,Hz to follow desired joint position. We carefully select motions that are safe for deployment, using this subset for training and deployment of MuGen.  
Our deployment codebase builds upon unitree\_rl\_gym\cite{unitreerobotics_unitree_rl_gym_2025}.
Since the default standing pose of the robot differs significantly from the first frame of the reference motion, we perform linear interpolation at the beginning to prevent large impulsive forces that could destabilize or damage the hardware. Results of real-world deployments are illustrated in Fig.~\ref{fig:teaser} (e-f). 

\section{Conclusion}
\label{sec:conclusion}

In this paper, we present a generative locomotion framework for humanoid robots based on a VQ-VAE skill embedding architecture. Our method distills a diverse skillset of locomotion skills from unstructured human motion data and produces a deployable student policy capable of real-time motion generation. The resulting controller can robustly track unseen motions and generate human-like behaviors without the need for handcrafted reward design.

\section{Limitations}
While our framework demonstrates the potential of generative models for humanoid locomotion and successfully builds what is, to the best of our knowledge, one of the first generative controllers for humanoid robots, several real-world constraints currently limit its scalability:

\begin{itemize}

\item \textbf{Difficulty with Standing Still.} We observe that our generative policy struggles to maintain a stable standing pose. This behavior arises because the VQ codebook was primarily trained on motion-rich sequences, and the quantized embeddings corresponding to static or quasi-static poses are underrepresented or ambiguous. As a result, the policy tends to alternate between similar but distinct embeddings, resulting in oscillatory motions like shuffling or pacing.

    \item \textbf{Inaccurate World Model.} Our learned dynamics model is a simple MLP and does not explicitly account for environmental interactions. While we observe that optimization is effective as long as gradients from the world model are directionally informative, its limited accuracy ultimately caps the performance of the teacher policy, especially for precise tracking.

\item \textbf{Limited Sim-to-Real Adaptation.} Our deployment relies solely on domain randomization to bridge the sim-to-real gap and stable deployment remains a challenge. A more robust approach would involve fine-tuning the world model using real-world rollouts and re-distilling a student policy accordingly.

\end{itemize}

\noindent\textbf{Future Work.} Potential directions include replacing MoCap with vision-based or onboard pose estimation, enhancing the world model with structure or uncertainty modeling, and scaling the skill codebook to encompass more diverse behaviors and terrains. Additionally, we envision integrating natural language interfaces—e.g., a GPT-style model mapping text prompts to skill sequences—for building a prompt-driven humanoid motion generation system.

\section*{ACKNOWLEDGMENT}
We thank Ruoyu Lin, Shiyi Xu, Yutong Liang, Liyuan Wang, Xurong Lu, Kunqi Xu, for help with deploying on G1 hardware. 
We thank Tao Huang for always being there to help with any problem and answer any question. We thank Yulong Zhang for invaluable feedback and for helping us use precise terminology throughout this work.
We thank Unitree Robotics for help with G1 support.

\bibliographystyle{IEEEtran}
\bibliography{example}

\input{sections/7_appendix}

\end{document}

%% file: sections/1_introduction.tex
\section{Introduction}
Humanoid robots have achieved significant advancements in bipedal locomotion and dynamic maneuvers (e.g., running, jumping, getting up, acrobatics) through deep reinforcement learning (RL) frameworks. However, careful task-specific reward shaping and tuning of RL algorithms are typically required for these tasks, which is often challenging and not generalizable across tasks. By leveraging the morphological similarity between humanoid robots and humans, along with the increasing availability of human motion performance datasets, imitating human performance has emerged as a promising approach for humanoid robots~\cite{Escontrela22arXiv_AMP_in_real, wang2023amp, ji2024exbody2, ma2025styleloco}. This approach alleviates the burden of precise reward shaping and enables robots to leverage the control strategies embedded in human motion to produce agile and expressive behaviors.

Existing methods for motion imitation typically train policies to take reference kinematic trajectories as part of the input and compute actions that control the robot to follow these trajectories. A kinematic motion generation module may also be employed to generate motions based on task goals, producing new reference trajectories that are then fed to the policy for tracking~\cite{ji2024exbody2}. However, this paradigm has two main limitations. First, naive tracking policies cannot effectively capture the structure of the action space under the kinematic and dynamic constraints of the robot, and they often struggle to extract distinctive motion patterns from large-scale datasets and reuse them in downstream tasks. Second, the separation of motion generation and robot control cannot guarantee the dynamic reliability of the generated trajectories. Even if the trajectories are kinematically plausible, the tracking policy may be unable to reproduce them, leading to suboptimal results or even failures.

In this paper, we address these fundamental limitations by learning a generative control system, \textbf{MuGen}, from a large human motion dataset with various locomotion performances. MuGen learns a structured latent space for motion skills and actions directly from physical simulation, ensuring that the skill embeddings are consistent with the kinematic and dynamic constraints of the robot. By projecting kinematic trajectories into this latent space or generating new trajectories from it, MuGen can achieve robust tracking on unseen motions and generation of new motions within the capabilities of the robot. 

We employ vector-quantized variational autoencoders (VQ-VAE) as the generative model, with the goal of reconstructing input motions from discrete latent codes. To achieve effective training, we draw inspiration from recent work in physics-based motion control~\cite{yao2022controlvae,yao2024moconvq} and adopt a model-based reinforcement learning (MBRL) approach, in which a differentiable dynamics model is trained to bridge the gap between kinematic references and the actions predicted by the policy. To further facilitate training, we adopt a two-stage training strategy. First, a teacher generative policy is learned using privileged information, and then the policy and skill embeddings are distilled into a deployable student policy. To this end, we develop a novel DAgger~\cite{dagger}-style distillation strategy with a progressive scheduling mechanism.

We evaluate MuGen on a diverse set of locomotion tasks and validate its performance in robust tracking and the generation of new motions on a Unitree G1 humanoid robot. Our approach significantly outperforms baseline methods in terms of tracking accuracy and robustness when trained on a small set of human motions and generalizes to unseen motions. We also demonstrate that MuGen can generate robust control of new motions using the latent embeddings without requiring reference trajectories.

%% file: sections/7_appendix.tex
\appendices

\section{Dataset}
\label{sec:dataset}

\begin{figure*}[h]
    \centering
    \includegraphics[width=1.0\linewidth]{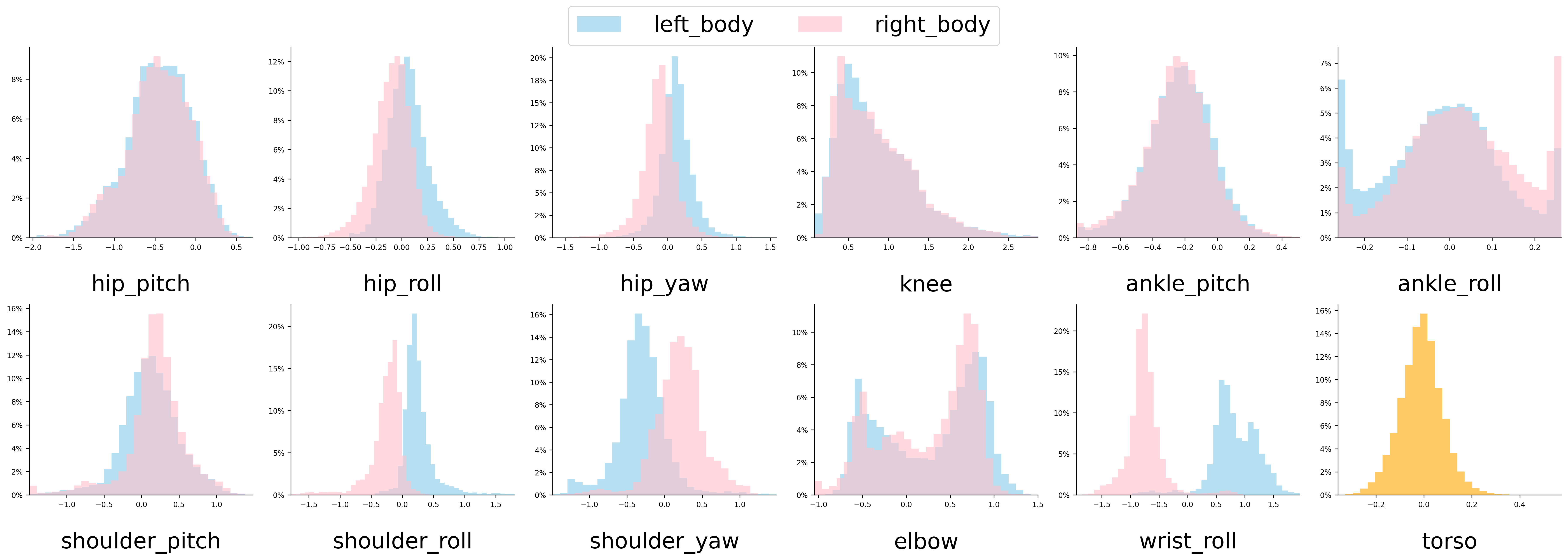}
    \caption{Data Distribution}
    \label{fig:data}
\end{figure*}

We provide here additional details on our locomotion data usage. We source our reference motions from the LaFAN1~\cite{harvey2020robustLafanYHY}, using its version retargeted to the G1 robot’s kinematic model as provided by Unitree~\cite{unitreerobotics_huggingface2025}. Table \ref{tab:dataset} details the composition of our dataset, which comprises walking and running sequences as well as the transitions between these two gaits. Figure~\ref{fig:data} demonstrates the data distribution of our selected subset: here “data distribution” refers to the histograms of joint‐angle values counted across all frames. A striking feature is that several key joints (notably the ankle, elbow, and knee) exhibit strongly non-Gaussian, often multimodal distributions rather than the unimodal, symmetric shapes one would expect under a normal assumption. This observation motivated our decision to forgo a VAE-based latent model, which typically assumes Gaussian priors in favor of a more flexible representation capable of capturing arbitrary, multimodal density profiles. We further selected some motions that do not involve self-collision and use this small set to conduct ablation study, including walking, crouching, turning around, and pacing. Table \ref{tab:dataset2} details the composition of our dataset for ablation study. We manually designed two reference motion sequences: one for walking straight forward and one for crouching walk straight forward. Both sequences were created by looping a single gait cycle and have been deployed on the real robot, with the walking-straight-forward sequence also used in an ablation study.

\begin{table}[h]
\centering
\caption{Composition of Used LaFAN1 Dataset}
\begin{tabular}{lrr}
\toprule
Name & Frames & Seconds \\
\midrule
walk2\_subject4 & 7146 & 238.2 \\
walk2\_subject3 & 7146 & 238.2 \\
walk3\_subject2 & 7399 & 246.63 \\
walk1\_subject5 & 7840 & 261.33 \\
walk2\_subject1 & 7146 & 238.2 \\
walk1\_subject1 & 7840 & 261.33 \\
walk3\_subject5 & 7399 & 246.63 \\
walk4\_subject1 & 4918 & 163.93 \\
walk1\_subject2 & 7840 & 261.33 \\
walk3\_subject3 & 7399 & 246.63 \\
\midrule
Sum Walk & 72073 & 2402.43 \\
\midrule
run2\_subject4 & 7345 & 244.83 \\
run1\_subject5 & 7135 & 237.83 \\
run2\_subject1 & 7345 & 244.83 \\
run1\_subject2 & 7135 & 237.83 \\
\midrule
Sum Run & 28960 & 965.33 \\
\bottomrule
\end{tabular}
\label{tab:dataset}
\end{table}

\begin{table}[h]
\centering
\caption{Composition of Ablation Dataset}
\begin{tabular}{lrr}
\toprule
Name & Frames & Seconds \\
\midrule
walk1\_subject1\_100\_1170 & 1070 & 35.67 \\
walk1\_subject5\_5740\_6830 & 1090 & 36.3 \\
walk1\_subject5\_80\_1715 & 1635 & 54.5 \\
walk1\_subject1\_2460\_3680 & 1220 & 40.67 \\
walk3\_subject5\_320\_1410 & 1090 & 36.3 \\
\midrule
Sum Train & 6105 & 203.5 \\
\midrule
walk1\_subject1\_2480\_2591 & 111 & 3.7 \\
walk1\_subject1\_2657\_3117 & 460 & 15.33 \\
walk1\_subject1\_5344\_5772 & 428 & 14.27 \\
walk\_straight(hand crafted) & 1590 & 53 \\
\midrule
Sum Test & 2589 & 86.3 \\
\bottomrule
\end{tabular}
\label{tab:dataset2}
\end{table}

\section{Observation Space}\label{app:obs_space}

The observation space of MuGen consists of two components: state and goal. We adopt two distinct coordinate spaces to represent information: global space and local space. 
For world model training, skill embedding learning, and teacher policy training, all calculations are performed in the global space. In contrast, student policy learning operates entirely in the local space.  
We maintain two separate state history buffers to store information in different spaces, and teacher and student policies can observe different numbers of history frames. Notably, we only feed the goal of the current frame to the policy, as providing goal history may degrade generative capability. Generally, observation at timestep $t$ is given by 
\begin{equation*}
    o_t=\{s_{t-H},\dots, s_{t},g_t\},
\end{equation*}
where $H$ means number of observed history frames, $g_t$ means tracking target $m_t$ in tracking policy while None in random sample mode.
\paragraph{Global Space} 
The humanoid is modeled as articulated rigid bodies with a floating root.  
A global state is defined as
\[
s^{\text{full}} = \{x_i, q_i, v_i, \omega_i\}_{i\in{B}},
\]
where \(B\) denotes the set of links including the floating root.  
We express it in the root heading frame as
\[
\bar{s}^{\text{full}} = \{\bar{x}_i, \bar{q}_i, \bar{v}_i, \bar{\omega}_i\},
\]
with the transformation
\[
(\bar{x}_i, \bar{q}_i, \bar{v}_i, \bar{\omega}_i) = q_0^{-1} \otimes (x_i - x_0, q_i, v_i, \omega_i).
\]
Rotations are represented using the 6D continuous format from \cite{zhou2019continuity}. After this transformation, at each timestep, the state \((\bar{x}_i, \bar{q}_i, \bar{v}_i, \bar{\omega}_i)\) has shape \((N_{\text{body}}, 3+6+3+3)\), corresponding to body position, body orientation (6D), linear velocity, and angular velocity.  
States and reference motions in global space encompass privileged information that is typically inaccessible to onboard sensing.

\paragraph{Local Space}
The local observation space reflects the sensing limitations of real robots, where privileged information such as root height and global velocity is inaccessible. A local state is defined as
\begin{equation*}
    s^{\text{proprioception}} = \{g, w\}\cup\{q_j, dq_j\}_{j\in{J}},
\end{equation*}
where \(J\) denotes the set of activated joints. Available observations include projected gravity $g$, body angular velocity $w$, and per-joint positions $q$ and velocities $dq$ relative to their parent joints.  
Combining these components, the local state for a single frame has shape \((3+3+N_{\text{joint}}\times2,)\).  
States and reference motions represented in local space contain only information realistically obtainable by onboard sensors.

\section{Domain Randomization}
\label{sec:domain_randomization}

To improve the robustness of the trained policy, we employ domain randomization to simulate several kinds of random noises that may occur while deploying in the real world. 
The terms used for randomization, along with their descriptions and ranges, are listed in Table\ref{tab:domain}.

\begin{table*}[!ht]
  \centering
  \caption{Randomization Terms, Description, and Ranges}
  \begin{tabularx}{\textwidth}{@{}l >{\raggedright\arraybackslash}X c@{}}
    \toprule[1.5pt]
    Term & Description & Ranges \\
    \midrule[1.5pt]
    Friction coefficient $(-)$ & Random friction coefficients applied to robot's links & $[0.60, 1.00]$ \\
    Torso payload mass $(Kg)$ & Additional random mass attached to the torso links & $[-2.00, 2.00]$ \\
    $K_p$ $(-)$ & Random scaling factors applied to the proportional gain ($K_p$) of the robot's joints & $[0.90, 1.10]$ \\
    $K_d$ $(-)$ & Random scaling factors applied to the derivative gain ($K_d$) of the robot's joints & $[0.90, 1.10]$ \\
    Dof pos obs $(rad)$& Random dof velocity added to the observation of joint positions & $[-0.05, 0.05]$ \\
    Dof vel obs $(rad/s)$& Random dof velocity added to the observation of joint velocities & $[-0.50, 0.50]$  \\
    Ang vel obs $(rad/s)$& Random dof velocity added to the observation of body angular velocities & $[-0.50, 0.50]$ \\
    Gravity obs $(m/s^2)$& Random dof velocity added to the observation of gravities projected to robot's body frame & $[-0.05, 0.05]$ \\
    \bottomrule[1.5pt]
  \end{tabularx}
  \label{tab:domain}
\end{table*}

\begin{table*}[!ht]
  \centering
  \caption{Hyperparameters, Description, and Value}
  \begin{tabularx}{\textwidth}{@{}l >{\raggedright\arraybackslash}X c@{}}
    \toprule[1.5pt]
    Term & Description & Value \\
    \midrule[1.5pt]
    num\_envs & number of parallel environment & 1024 \\
    $T$ & maximum episode length in single rollout trajectory & 1500 \\
    $T_m$ & length of sampled reference motion clip & 120\\
    FPS & Frames Per Second of simulation, reference motion, and control signal & 30 \\
    Substep & substep of simulation within each control step & 6 \\
    buff\_len & maximum length of replay buffer & 256 \\
    $h_0$ & height threshold for termination: episode terminates when any non-foot link’s vertical position falls below this value & 0.2 m \\
    $BS^\wm$ & Batch Size for training world model & 512\\
    $BS^\mathcal{T}$  & Batch Size for training teacher policy & 1024\\
    $BS^\mathcal{S}$  & Batch Size for training student policy & 1024\\
    $l^\wm$ & replay clip length for training world model & 24\\
    $l^\mathcal{T}$ & replay clip length for training teacher policy & 24\\
    $l^\mathcal{S}$ & replay clip length for training student policy & 32\\
    lr & learning rate & 2e-4\\
    $\gamma$ & discount factor & 1\\
    $W_{\wm}$ & weight of state component(position, rotation, velocity, angular) loss for training world model & [2,1,10,5]\\
    $W_{T}$ & weight of state component(position, rotation, velocity, angular) loss for training teacher policy & [2,1,0.5/3,0.5/3]*0.1\\
     $\beta_1$ &  weight of commitment loss & 0.05\\
     $\beta_2$ & weight of smooth action loss & 0.01\\
     $\beta_3$ & weight of action L2-norm penalty loss & 0.001\\
     $\beta_4$ &  weight of align  loss & 1\\
    \bottomrule[1.5pt]
  \end{tabularx}
  \label{tab:hyper}
\end{table*}

\section{Pseudo Code}
We show our pseudo-code for training MuGen in \alg\ref{alg:training_mugen} and \alg\ref{alg:training_modules}. 
MuGen is trained through a staged pipeline comprising three core phases and one optional refinement step. First, during \emph{pretraining} ($\text{epoch}<ms_1$), the teacher policy explores the motion space and builds the latent skillset in the VQ codebook, jointly optimized with the world model. 
Second, in the \emph{warming-up} phase ($ms_1<\text{epoch}<ms_2$), both teacher and student policies are trained, but only the teacher is used to generate rollouts. This helps the student gain a stable imitation target. Third, in the \emph{distillation} phase ($ms_2<\text{epoch}<ms_3$), control is gradually handed over to the student policy using a linear annealing strategy that mixes teacher and student actions within each trajectory, enabling smooth policy transition without full dependence on DAgger-style aggregation.
An optional \emph{post-training} phase ($\text{epoch}>ms_3$) further adapts the student policy to selected reference motions for improved real-world robustness. The full procedure is detailed in \alg\ref{alg:training_mugen}. In each training step, we update the modules (world model, teacher policy, and student policy) by sampling trajectory clips and computing the corresponding objective functions as described in \Sec\ref{sec:method}. Since we operate in two distinct information spaces—global and local, we store trajectory data for both spaces separately. During the joint training of the world model, codebook, and teacher policy, we sample only global space trajectories from the replay buffer. In contrast, during the distillation phase, we sample both global and local trajectories, as the student policy requires demonstrations from the teacher, which operates in the global space.

\begin{algorithm}
    \SetAlgoLined    
    \DontPrintSemicolon
    
      \SetKwProg{Train}{Function}{:}{end}
      \Train{ \textnormal{\textbf{Train}$($ $)$  }}
      {
        Hyperparameters: milestones $ms=[ms_1, ms_2, ms_3]$, max expisode length $T$, reference motion clip length$T_m$, trajectory clip length $l_\wm$, $l_\mathcal{T}$, $l_\mathcal{S}$ \;
        Initialize: teacher policy $\policy^{\text{T}}={\encoder^{\text{T}}, \codebook, \decoder^{\text{T}}}$, student policy $\policy^{\text{S}}={\encoder^{\text{S}}, \decoder^{\text{S}}}$, shared codebook $\codebook$, world model $\wm$, trajectory buffer $\simBuffer \gets \varnothing  $, epoch counter $e \gets 0$ \;
        \While{\textnormal{training}}{
            \tcp{collect simulation trajectories}
            Remove the oldest $N_{B}'$ simulation tuples from $\simBuffer$ \;
            \While{$|\simBuffer| < N_{B}$}{
                Select $\tilde{\tau}=\{\tilde{\refm}_0, \tilde{\refm}_1, \dots{},\tilde{\refm}_T\}$ from D \;
                ${\tau} \gets$ \textbf{TakeOver}($\tilde{\tau}$, $\tilde{\refm}_0$, $T$, $\policy^{\text{T}}$, $\policy^{\text{S}}$, $\codebook$) \;
                Store ${\tau}$ and $\tilde{\tau}$ in $\simBuffer$\;
            }
            \If{$e<ms_2$}
            {
                \textbf{TrainWorldModel}($\wm$, $l$, $\simBuffer$) \;
                \textbf{TrainTeacher}($\wm$, $\policy^{\text{T}}$, $\codebook$, $l$, \simBuffer) \;
            }
            \If{$e>ms_1$}
            {
                \textbf{TrainStudent}($\policy^{\text{T}}$, $\policy^{\text{S}}$, $\codebook$, $l$, \simBuffer) \;
            }
        }
      }          
      
      \SetKwProg{TakeOver}{Function}{:}{end}
      \TakeOver{ \textnormal{\textbf{TakeOver}$($ $\tilde{\tau}$, $\stt_0$, $T$,$\policy^{\text{T}}$, $\policy^{\text{S}}$, $\codebook$ $)$} }
      {    
        $t \gets 0$\;
        \While{\textnormal{not terminated $and$ $t < T$}}{

            \uIf { $e<ms_2$ }
            {
                $P\gets 1$ \; 
            }
            \uElseIf { $e>ms_3$ }
            {
                $P\gets 0$ \; 
            }
            \Else 
            {
                $P\gets 1 - (e-ms_2) / (ms_3-ms_2)$ \; 
            }
            Randomly select policy $\policy$ with probability $p(\policy^{\text{T}})=P$ and $p(\policy^{\text{S}})=1-P$\;
            Extract $\tilde{\refm}^t$ from $\tilde{\tau}$\;
            Sample $\act_t \sim{} \policy(\act_t|s_t, \tilde{\refm}_t, \codebook)$ by \textbf{Policy}($\act_t|s_t, \tilde{\refm}_t, \codebook$)\;
            
            ${\stt}_{t+1} \leftarrow$ Simulate$({\stt}_t$, $\act_t)$ \;
            
            $t \leftarrow t+1$\;

            \If {run out of reference motion clip}
            {
            Resample $\tilde{\tau}=\{\tilde{\refm}_0, \tilde{\refm}_1, \dots{},\tilde{\refm}^T_m\}$ from D \;
            }
        }
        Return ${\tau}=\{{\stt}_0, \act_0, {\stt}_1, \act_1, \dots, \}$\; 
      }  
      
    \SetKwProg{Policy}{Function}{:}{end}
      \Policy{ \textnormal{\textbf{Policy}$($ $\act_t|s_t, \tilde{\refm}_t, \codebook$ $)$  }}
      {
        Sample $\latent_t \sim{} \encoder(\latent_t|{\stt}_t,\tilde{\refm}_{\text{T}})$\;
        Quantize $\hat{\latent}_t \sim \codebook(\latent_t)$\;
        Sample ${\act}_t \sim{} \decoder({\act}_t|{\stt}_t,\hat{\latent}_t)$ \;
      }
      
    \caption{Train MuGen}
    \label{alg:training_mugen}
\end{algorithm}

\begin{algorithm}
  \SetAlgoLined    
  \DontPrintSemicolon
  
      \SetKwProg{TrainWorldModel}{Function}{:}{end}
      \TrainWorldModel{ \textnormal{\textbf{TrainWorldModel}$($ $\wm$, $T$, $\simBuffer{}$ $)$  }}
      {
        Sample ${\tau}^*=\{{\stt}_0, {\act}_0, {\stt}_1, {\act}_1, \dots, , {\stt}_T, {\act}_T\}$ from 
        $\simBuffer$, ignore $\tilde{\tau}^*$ \;
        $\bar{\stt}_0\gets\stt_0$\;
        \For{$t\gets0$ \KwTo $T$}{
            $\bar{\stt}_{t+1} \leftarrow \wm(\bar{\stt}_t,\act_t)$ \;
        }
        Cache ${\tau}=\{{\stt}_0, {\stt}_1, \dots, {\stt}_T\}$\; 
        Cache $\bar{\tau}=\{\bar{\stt}_0, \bar{\stt}_1, \dots, \bar{\stt}_T\}$\; 
        $\loss \gets \loss_{\text{w}}(\bar{\tau},{\tau}^*) $ \;
        Update $\wm$ with $\loss{}_{\text{w}}$
      }

      \SetKwProg{TrainTeacher}{Function}{:}{end}
      \TrainTeacher{ \textnormal{\textbf{TrainTeacher}$($ $\wm$, $\policy$, $\codebook$, $T$, \simBuffer $)$} }
      { 
        Sample ${\tau}^*=\{{\stt}_0, {\act}_0, {\stt}_1, {\act}_1, \dots, , {\stt}_T, {\act}_T\}$ and $\tilde{\tau}^*=\{\tilde{\refm}_0, \tilde{\refm}_1, \dots{},\tilde{\refm}_T\}$ from $\simBuffer$\;
        $\bar{\stt}_0\gets\stt_0$\;
        \For{$t\gets0$ \KwTo $T$}{
            Sample $\bar{\act}_t \sim \policy(\act_t|\bar{\stt}_t, \tilde{\refm}_t, \codebook)$\;
            
            $\bar{\stt}_{t+1} \leftarrow \wm(\bar{\stt}_t,\bar{\act}_t)$ \;
            
        }
        Cache $\bar{\tau}=\{\bar{\stt}_0, \bar{\act}_0, \bar{\stt}_1, \bar{\act}_1, \dots, \bar{\stt}_T, \bar{\act}_T\}$\; 
        $\loss \gets \loss_{\text{rec}}(\bar{\tau},\tilde{\tau}^*)  + \beta \loss_{kl}(\bar{\tau}) +  \loss_{\text{act}}(\bar{\tau})$ \;

Update $\policy^{\text{T}}, \codebook$ with $\loss$
    }

    \SetKwProg{TrainStudent}{Function}{:}{end}
      \TrainStudent{ \textnormal{\textbf{TrainStudent}$($ $\policy^{\text{T}}$, $\policy^{\text{S}}$, $\codebook$, $T$, \simBuffer $)$} }
      { 
        Sample ${\tau}^*=\{{\stt}^0, {\act}^0, {\stt}^1, {\act}^1, \dots, , {\stt}^T, {\act}^T\}$ and $\tilde{\tau}^*=\{\tilde{\refm}^0, \tilde{\refm}^1, \dots{},\tilde{\refm}^T\}$ from $\simBuffer$\;
        \For{$t\gets0$ \KwTo $T$}{
            Sample $\bar{\act}^{\text{T}}_t \sim \policy^{\text{T}}(\act^t|{\stt}_t^{\text{T}}, \tilde{\refm}_t, \codebook)$\;
            Sample $\bar{\act}^{\text{S}}_t \sim \policy^{\text{S}}(\act_t|{\stt}_t^{\text{S}}, \tilde{\refm}_t, \codebook)$\;
        }
        Cache $\bar{\tau}^{\text{T}}=\{\bar{\latent}^{\text{T}}_0, \bar{\act}^{\text{T}}_0, \bar{\latent}^{\text{T}}_1, \bar{\act}^{\text{T}}_1, \dots, \bar{\latent}^{\text{T}}_T, \bar{\act}^{\text{T}}_T\}$\; 
        Cache $\bar{\tau}^{\text{S}}=\{\bar{\latent}^{\text{S}}_0, \bar{\act}^{\text{S}}_0, \bar{\latent}^{\text{S}}_1, \bar{\act}^{\text{S}}_1, \dots, \bar{\latent}^{\text{S}}_T, \bar{\act}^{\text{S}}_T\}$\; 
        $\loss \gets \loss_{\text{rec}}(\bar{\tau}^{\text{S}},\bar{\tau}^{\text{T}})$ \;
        
        Update $\policy^{\text{S}}$ with $\loss$\;
    }      
\caption{Train Step for world model, teacher policy, and student policy}
\label{alg:training_modules}
\end{algorithm}

\section{Network Structure}\label{app:Network}

Our architecture consists of four main components: a shared vector-quantised (VQ) bottleneck \(\codebook\), an encoder \(\encoder\) to process embedded information, a policy decoder \(\decoder\) that outputs actions from the discrete quantized latent code, and a world model \(\wm\) serving as a differentiable simulator. Together, \(\embedding\) and \(\encoder\) form the latent extractor, as illustrated in \fig\ref{fig:pipeline}.
The teacher and student policies share almost identical network architectures, differing only slightly in input dimensionality.

We first flatten the input states and pass them through an embedding layer \(\embedding_s\). The target goal for a single frame is separately embedded via \(\embedding_g\). These two embeddings are concatenated and then fed into the encoder \(\encoder\) to produce a latent representation in a continuous latent space. The decoder \(\decoder\) then processes the vector-quantized skill embedding searched from codebook $\codebook$ to produce a latent action, which is further passed through two small linear layers to predict the mean (\(\mu\)) and standard deviation (\(\sigma\)) of the action distribution.

\section{Ablation Baselines}\label{sec:ablation_baseline}
Here we give more detailed information of our baselines in \Tab\ref{tab:baseline}.  

\textbf{VQ-based w/o history:} This model shares the exact architecture with our full MuGen model, except that it does not use historical states in the observation (i.e., \#HT = \#HS = 0). This ablation helps evaluate the impact of temporal context in the input representation.

\textbf{VAE-based w/o history:} In this variant, we remove the vector quantization codebook and instead apply a Kullback-Leibler (KL) divergence loss to the continuous latent variable $z$, encouraging it to follow a standard normal distribution. All other components remain identical to the 'VQ-based w/o history' model.

\textbf{MLP baseline:} As a baseline, we use a standard MLP-based model with the same architecture as the 'VAE-based w/o history' variant, but without any KL regularization. 

\section{Visualization}

To visualize the kinematic behavior of our system, we construct a high-fidelity rendering pipeline based on Blender 4.4. The pipeline consists of three main stages: (1) \emph{Motion export}: the simulated trajectory, including link positions and orientations, is exported from Isaac Gym at 30\,Hz and saved in \texttt{.npy} format; (2) \emph{Scene construction}: the robot’s 3D geometry in \texttt{.stl} format is imported into Blender, where the scene is assembled based on XML descriptors. Materials, lighting, and camera parameters are configured for photorealism; (3) \emph{Animation scripting}: the exported motion is mapped onto Blender’s rigged model via Python scripts, where transformation matrices are applied frame-by-frame and converted into animation keyframes. 
This pipeline allows us to generate consistent visualizations for simulated trajectories, enabling direct qualitative comparison and clearer demonstration of motion quality.